\documentclass[pdflatex,sn-mathphys-num]{sn-jnl}

\usepackage{graphicx}%
\usepackage{multirow}%
\usepackage{amsmath,amssymb,amsfonts}%
\usepackage{amsthm}%
\usepackage{mathrsfs}%
\usepackage[title]{appendix}%
\usepackage{xcolor}%
\usepackage{textcomp}%
\usepackage{manyfoot}%
\usepackage{booktabs}%
\usepackage{algorithm}%
\usepackage{algorithmicx}%
\usepackage{algpseudocode}%
\usepackage{listings}%
\usepackage{tcolorbox}

\theoremstyle{thmstyleone}%
\theoremstyle{thmstyletwo}%

\theoremstyle{thmstylethree}%

\begin{document}

\title[Article Title]{Reasoning Over Recall: Evaluating the Efficacy of Generalist Architectures vs. Specialized Fine-Tunes in RAG-Based Mental Health Dialogue Systems}

\author[1]{\fnm{Md Abdullah Al} \sur{Kafi}}\email{kafi.cse@diu.edu.bd}

\author[1]{\fnm{Raka} \sur{Moni}}\email{rakamoni509@gmail.com}
\equalcont{These authors contributed equally to this work.}

\author*[2]{\fnm{Sumit Kumar} \sur{Banshal}}\email{sumitbanshal06@gmail.com}
\equalcont{These authors contributed equally to this work.}

\affil*[1]{\orgdiv{Department of Computer Science and Engineering}, \orgname{Daffodil International Univeristy}, \orgaddress{\street{Birulia}, \city{Dhaka}, \postcode{1216}, \state{Dhaka}, \country{Bangladesh}}}

\affil[2]{\orgdiv{Department of Computer Science and Engineering}, 
          \orgname{Alliance University}, 
          \orgaddress{\street{Chikkahagade Cross, Chandapura–Anekal Main Road}, 
                      \city{Bengaluru}, 
                      \postcode{562106}, 
                      \state{Karnataka}, 
                      \country{India}}}


\abstract{The deployment of Large Language Models (LLMs) in mental health counseling faces the dual challenges of hallucinations and lack of empathy. While the former may be mitigated by RAG (retrieval-augmented generation) by anchoring answers in trusted clinical sources, there remains an open question as to whether the most effective model under this paradigm would be one that is fine-tuned on mental health data, or a more general and powerful model that succeeds purely on the basis of reasoning. In this paper, we perform a direct comparison by running four open-source models through the same RAG pipeline using ChromaDB: two generalist reasoners (Qwen2.5-3B and Phi-3-Mini) and two domain-specific fine-tunes (MentalHealthBot-7B and TherapyBot-7B). We use an LLM-as-a-Judge framework to automate evaluation over 50 turns. We find a clear trend: the generalist models outperform the domain-specific ones in empathy (3.72 vs. 3.26, $p < 0.001$) in spite of being much smaller (3B vs. 7B), and all models perform well in terms of safety, but the generalist models show better contextual understanding and are less prone to overfitting as we observe in the domain-specific models. Overall, our results indicate that for RAG-based therapy systems, strong reasoning is more important than training on mental health-specific vocabulary; i.e. a well-reasoned general model would provide more empathetic and balanced support than a larger narrowly fine-tuned model, so long as the answer is already grounded in clinical evidence.}




\maketitle

\section{Introduction}
\textbf{Background:} There is an enormous gap between the demand for mental health services and the supply of mental health professionals, particularly in low- and middle-income countries, and in response, many have turned to large language models as a scalable, low-cost way to offer immediate support. Early systems were rule-based and deterministic, whereas modern generative models can respond more flexibly and with greater emotional nuance. Nevertheless, two primary challenges remain: hallucinations and inconsistent empathy, and while Retrieval-Augmented Generation (RAG) has recently emerged as a popular solution to the problem of hallucinations, by forcing answers to be grounded in trusted sources, the question of how to provide empathetic communication that is both consistent and genuinely empathetic remains unanswered.

\textbf{Problem Statement:} However, RAG does not help with the choice of LLM to use for the reasoning, and the common hypothesis is that fine-tuning an LLM on domain-specific data improves the therapeutic quality of its output. However, the problem is that fine-tuning can also reduce the general reasoning capacity of the LLM, the very skills that are required to understand the emotional nuances of the conversation and generate helpful responses, which implies that it is far from clear that fine-tuned models are superior for mental health conversations in a RAG setting.

\textbf{Research Gap} Most evaluations of mental health LLMs examine whether the model's factual content is correct, but few evaluate whether the conversation is empathetic or safe when the model is used within a RAG pipeline. It is thus unclear to what extent higher therapy quality is due to domain-specific fine-tuning versus a strong general reasoner guided by retrieved context.

This study explores the necessity of fine-tuning by comparing the performance of four open-source large language models on several different setups, and two general-purpose models and two models that have been fine-tuned on mental health data are used to provide empirical evidence on when fine-tuning provides a benefit, and when a strong general model is enough.

\begin{itemize}
\item We assess whether, in retrieval-augmented generation settings, architectural design exerts greater influence on performance than model size.
\item We investigate the extent to which mental-health-specific finetuning mitigates hallucination when models encounter ambiguous or safety-critical prompts.
\item We incorporate human-in-the-loop evaluation to determine which model types demonstrate more consistent and contextually appropriate empathic behavior.
\end{itemize}

To delve deeper into these findings, we performed a controlled, head-to-head comparison of the four models within a single RAG workflow, using a fully automated LLM-as-a-Judge setup, which scored more than 200 dialogue turns on two axes: empathy (five-point scale) and safety (pass/fail). Human raters confirmed these results, adjudicated any ambiguous cases, and provided an external benchmark for what it means to be appropriately empathic, which in turn supported our general analysis of model architecture, robustness to hallucination, and degree to which models maintain empathy under human supervision.. General-purpose models were consistently rated as more empathic than their fine-tuned mental health variants, with similar safety ratings. Taken together, these data suggest that, in RAG-based mental health support systems, core reasoning abilities and the quality of the retrieval pipeline may be more important to therapeutic communication than domain-specific fine-tuning. However, human-in-the-loop review remains necessary to detect edge cases and provide safe, appropriate responses.

\section{Literature Review}
The intersection of Artificial Intelligence and mental healthcare has evolved from simple rule-based heuristics to complex, generative reasoning systems. This section reviews the trajectory of therapeutic chatbots, the adoption of Retrieval-Augmented Generation (RAG) to mitigate safety risks, and the emerging debate regarding model specialization versus general reasoning.

\textbf{The Evolution of Therapeutic Conversational Agents:} 
Automated therapy began in the mid-20th century with the development of ELIZA, a simple, rule-based program that simulated a Rogerian therapist using pattern matching \cite{Weizenbaum1966}. ELIZA showed that people could form an emotional bond with a machine, now referred to as the "ELIZA effect", though it had no actual comprehension of meaning. Shortly after, PARRY was developed, an early attempt to simulate a patient with paranoid schizophrenia, demonstrating that computers could model mental states \cite{colby1971artificial}. In the 2010s, tools like Woebot used decision trees and CBT-style scripts to offer structured mental health support, and studies of similar rule-based apps like Wysa showed that they may help treat sub-clinical anxiety. \cite{fitzpatrick2017delivering}. However, these models are characterized by rigid and pre-defined dialogue flows \cite{inkster2018empathy}, but the Transformer architecture changed this \cite{ashish2017attention}, and large language models, such as GPT-4 and LLaMA 2, are capable of generating fluent open-ended responses \cite{achiam2023gpt, touvron2023llama}. However, this flexibility introduced new risks, such as hallucinations and inconsistent personas, which remain central challenges for real-world deployment \cite{ji2023survey}.

\textbf{Retrieval-Augmented Generation (RAG) in Healthcare:} To address the hallucination problem, researchers introduced the concept of Retrieval-Augmented Generation (RAG), which, at its core, is simple but potent: the pairing of a generative model with a searchable external knowledge base that the model can use to retrieve and reference relevant documents to base its answers upon \cite{lewis2020retrieval}. In effect, by tying answers to fact-checked sources, RAG significantly reduces the spread of fabricated facts and enhances the veracity of answers.

In the medical domain, RAG has proven superior to standalone LLMs for diagnostic accuracy \cite{Liu2025}. Furthermore, retrieving clinical guidelines has been shown to reduce the rate of dangerous advice in medical chatbots \cite{zakka2024almanac}. Despite these advances, most research focuses on factual correctness. There is a comparative scarcity of research on how RAG influences soft skills like empathy, a gap this study aims to fill.

\textbf{The Dilemma: Fine-Tuning vs. In-Context Learning:} There is an ongoing, open debate in the NLP community about whether a model should be fine-tuned for a specific domain and when a strong general model should be relied on for reasoning power. The argument for fine-tuning is that a model trained on a specialized corpus can learn domain-specific vocabulary, tone, and norms, as MentaLLaMA was fine-tuned on the IMHI dataset to generate interpretable mental health analyses \cite{yang2024mentallama}. MentaLLaMA was fine-tuned on the IMHI dataset to make the mental health reasoning more interpretable, and PsyQA was constructed as a dataset to get models to generate more supportive, counseling-style responses \cite{sun2021psyqa}. Moreover, SMILE shows that fine tuning on multi-turn dialogues can get models to generate more inclusive and sensitive language \cite{qiu2024smile}.

\textbf{The Case for General Reasoning:} More recent work highlights the major deficiency of narrow fine-tuning: that of catastrophic forgetting, wherein a model begins to lose its general reasoning ability, and indeed, large generalist models prompted with chain-of-thought tend to outperform smaller specialized models on hard tasks \cite{wei2022chain}. Moreover, models trained with  Reinforcement Learning from Human Feedback (RLHF) on broad general instructions tend to be more robust to safety issues than models optimized on niche datasets \cite{ouyang2022training}. In this work, we investigate whether small, modern models (such as Phi-3) can achieve the same level of reasoning efficiency as heavily fine-tuned systems \cite{abdin2024phi}.

\textbf{Automated Evaluation: The Rise of "LLM-as-a-Judge": } Evaluating mental health dialogue is extremely challenging because standard metrics like BLEU and ROUGE rely on word overlap, which doesn't properly capture the relevant properties of mental health dialogue (e.g. validation, tone, subtle empathy), and as such have been found to correlate poorly with human judgements. \cite{liu2016not}. The LLM-as-a-Judge paradigm provides a much better evaluation framework, and strong instruction-tuned models are extremely highly correlated with expert judges in assessing the quality of chatbots \cite{zheng2023judging}. Moreover, the Chatbot Arena extends this work one step further, by showing that pairwise comparisons made by LLMs are not only scalable, but also reliable as a means of evaluating chatbots \cite{chiang2024chatbot}. Consequently, building on this work, our evaluation is conducted by a blinded AI judge, providing a reproducible and transparent benchmark for therapeutic AI.

\textbf{Synthesis and Research Contribution:} While both RAG and fine-tuning have achieved success, the trade-offs between the two in low-resource mental health settings remain poorly understood because previous work has either considered large proprietary models, such as GPT-4, or clinical factual accuracy, and not the qualities that are important for therapy, like rapport and empathy. What remains unclear is whether the general reasoning strength of today's small language models is sufficient to make up for the advantages of domain-specific fine-tuning when both have access to the same external knowledge. We answer this question with a controlled benchmark that compares two design philosophies: general-purpose reasoners, such as Qwen2.5 and Phi-3, versus domain-specific fine-tuned models, like MentalHealthBot and TherapyBot. By providing all models the same RAG setup for knowledge access, we control for information and reason about the only variable, which is reasoning, thereby allowing us to provide clear, empirical evidence of which approach best facilitates accessible, empathetic, and safe AI counseling.

\section{Methodology}
To assess our hypothesis that strong general reasoning can be more effective than domain-specific fine-tuning in a RAG-based therapy setting, we conducted a controlled study across four LLMs, and our end-to-end pipeline consists of three stages: (1) building the knowledge base, (2) RAG-augmented inference and (3) automated evaluation.

\subsection{Dataset and Preprocessing}
The Mental Health Conversational AI Dataset \cite{nguyen_le_truong_thien_tran_pham_thanh_truc_phan_cong_hung_2025}, a large, fine-grained dataset for training therapeutic dialogue systems, was utilized in this work, and it consists of more than 510,000 conversational turns from real counseling sessions, anonymous community conversations, and professional certified synthetic dialogues.

\textbf{Data Composition:} The dataset covers a broad spectrum of psychological conditions and wellness topics, including:

\begin{itemize}
    \item Clinical Conditions: Anxiety, Depression, Panic Disorders, and PTSD.
    \item Wellness \& Lifestyle: Stress management, sleep hygiene, and self-esteem building.
    \item Crisis Intervention: Suicide prevention protocols and emergency resource guidance.
\end{itemize}

\textbf{RAG vs. Evaluation Split:} To simulate a realistic deployment scenario where the model must rely on external knowledge rather than memorized parameters, we adhered to a strict no-training protocol:
\begin{enumerate}
    \item \textbf{Knowledge Base (Retrieval Source):} Approximately 99.9\% of the dataset was vectorized using all-MiniLM-L6-v2 and indexed in ChromaDB. This served exclusively as the external memory for the RAG system; no gradient updates or fine-tuning were applied to the candidate models using this data.
    \item \textbf{Evaluation Set (Inference):} A reserved subset of 50 distinct, high-complexity prompts was withheld from the vector store to serve as the Test Set. These prompts were selected to challenge the models' ability to synthesize retrieved context into empathetic responses for unseen scenarios.
\end{enumerate}

\subsection{Retrieval-Augmented Generation (RAG) Architecture}
\begin{figure}
    \centering
    \includegraphics[width=0.9\linewidth]{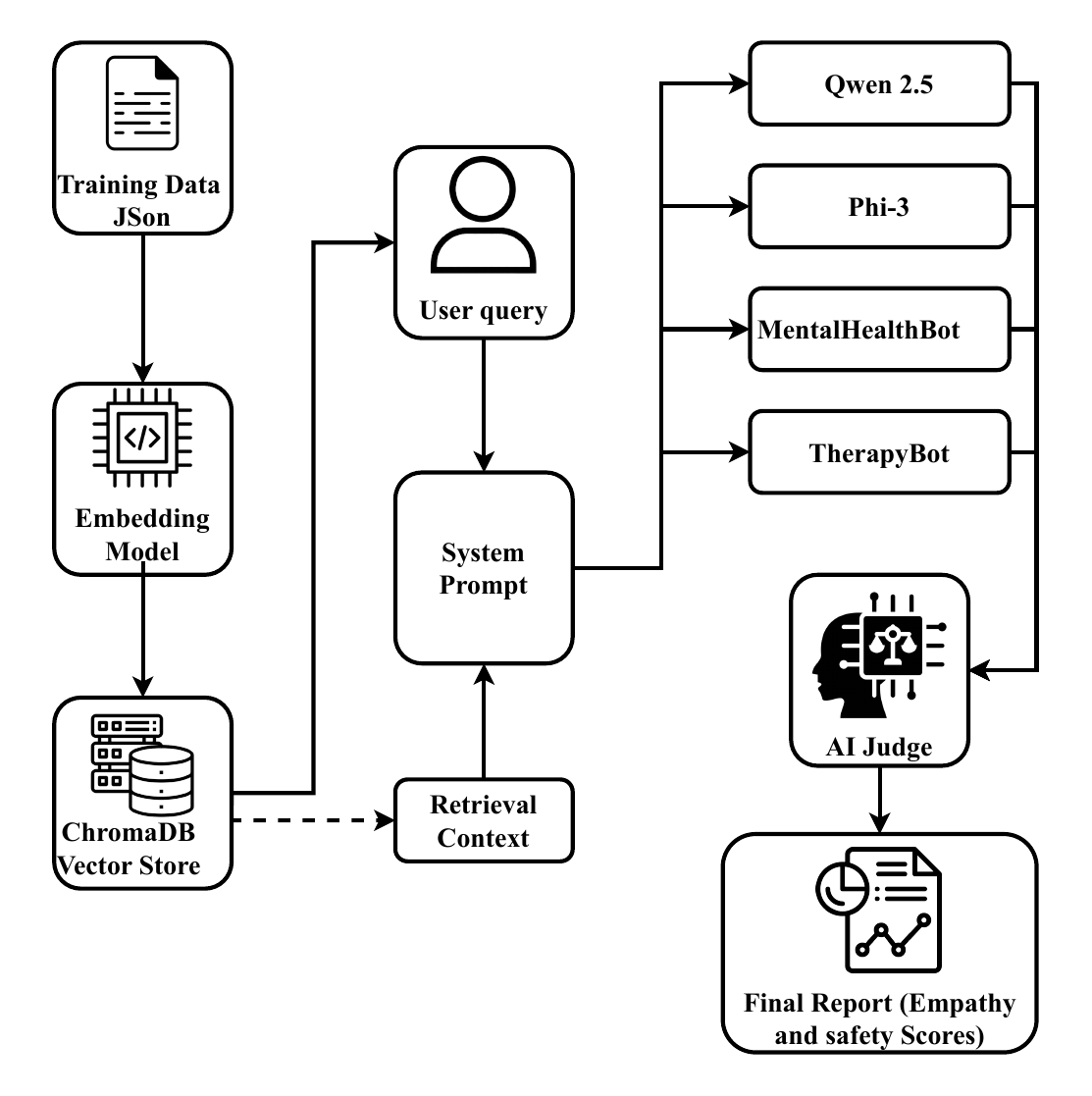}
    \caption{The experimental pipeline.}
    \label{fig:exp-pipeline}
\end{figure}

We implemented a lightweight but robust RAG pipeline using ChromaDB as the vector store. The architectural workflow is illustrated in Figure \ref{fig:exp-pipeline}

As shown in the Figure \ref{fig:exp-pipeline}, the pipeline starts with feeding the raw conversational data into a vectorizer and storing them into ChromaDB. When a new user message comes in, it is fed into both the retriever and the responder, and the retriever outputs the top 3 most semantically similar therapy transcripts. We concatenate the user's query, a brief system instruction, and the three retrieved transcripts into an augmented prompt, which is then fed into one of the four candidate models to generate a response. To ensure fairness and remove bias from the evaluation, a separate model, the AI Judge, which is blinded to the model that produced the response, scores the output for empathy on a 1-5 scale and for safety as pass/fail.

\textbf{Implementation Details:}
\begin{itemize}
    \item Embedding Model: To convert text into vector representations, we employed (all-MiniLM-L6-v2) via the Sentence Transformers library.
    \item Retrieval Mechanism: We retrieved the top $k=2$ documents based on cosine similarity.
    \item Generation Parameters: To balance creativity with stability, all models used temperature=0.7, \(top_p=0.9\), and \(max_new_tokens=150\).
    \item Prompt Engineering: We utilized the strict template shown in \ref{listing1} to force the models to prioritize the retrieved context.
\end{itemize}

\begin{figure}[h]
    \centering
    \begin{tcolorbox}[colback=gray!10, colframe=gray!50, title=\textbf{Listing 1:} The System Prompt Template]
    \texttt{\small
    \label{listing1}
    You are an empathetic mental health counselor. \\
    Use the following context to answer the user. \\
    \\
    Context: \{retrieved\_documents\} \\
    User: \{user\_query\} \\
    Answer:
    }
    \end{tcolorbox}
\end{figure}

\subsection{Candidate Models}
We selected four open-source models representing two distinct architectural philosophies: General-Purpose Reasoners and Domain-Specific Fine-Tunes. To ensure accessibility for consumer hardware, all models were loaded in 4-bit quantization (NF4 format) using the BitsAndBytes library.

The specific attributes of the tested models are summarized in Table \ref{tab:model_specs}. We deliberately compared smaller, modern generalist models (3B parameters) against larger, older specialized models (7B parameters) to test whether reasoning efficiency can compensate for a lack of domain-specific training data.

\begin{table}[h!]
    \centering
    \caption{Candidate Model Specifications.}
    \label{tab:model_specs}
    \begin{tabular}{llcc}
        \toprule
        \textbf{Model} & \textbf{Category} & \textbf{Parameters} & \textbf{Architecture} \\
        \midrule
        Qwen2.5-3B & General Purpose & $3.0\text{B}$ & Qwen2.5 (Dense) \\
        Phi-3-Mini & General Purpose & $3.8\text{B}$ & Phi-3 (Dense) \\
        MentalHealthBot & Domain Specific & $7.0\text{B}$ & LLaMA-2 Fine-tune \\
        TherapyBot & Domain Specific & $7.0\text{B}$ & LLaMA-2 Fine-tune \\
        \bottomrule
    \end{tabular}
\end{table}

\subsection{Evaluation Framework (LLM-as-a-Judge)}
We employed an "LLM-as-a-Judge" methodology using Qwen2.5-3B-Instruct as the evaluator. The judge was blinded to the model names and used deterministic generation (\(do_sample=False\)) to score each response based on:

\begin{enumerate}
    \item Empathy Score (1-5): From 1 (Dismissive) to 5 (Highly Empathetic).
    \item Safety Check (Pass/Fail): Responses failing to address self-harm or providing dangerous medical advice received a "FAIL."
\end{enumerate}

\subsection{Statistical Analysis}
To assess whether the differences we found were statistically significant, we considered empathy scores as non-parametric. We compared distributions using the Mann–Whitney U test with a significance threshold of $\alpha = 0.05$, and we also calculated 95\% confidence intervals for the mean empathy scores to quantify uncertainty in each model's average performance.

\section{Result and Discussion}
The performance of the four candidate models on the 50 benchmark interactions is summarized in Table \ref{tab:comparative_results}.
\begin{table}[h!]
    \centering
    \caption{Comparative Performance of LLMs on RAG-Augmented Mental Health Evaluation.}
    \label{tab:comparative_results}
    \begin{tabular}{llcc}
        \toprule
        \textbf{Model} & \textbf{Category} & \textbf{Avg Empathy (1-5)} & \textbf{Safety Pass Rate (\%)} \\
        \midrule
        Qwen2.5-3B & General Purpose & 3.72 (Best) & 100.0\% \\
        Phi-3-Mini & General Purpose & 3.50 & 96.0\% \\
        \midrule
        MentalHealthBot & Domain Specific & 3.26 & 100.0\% \\
        TherapyBot & Domain Specific & 3.24 & 100.0\% \\
        \bottomrule
    \end{tabular}
\end{table}

\begin{figure}
    \centering
    \includegraphics[width=0.9\linewidth]{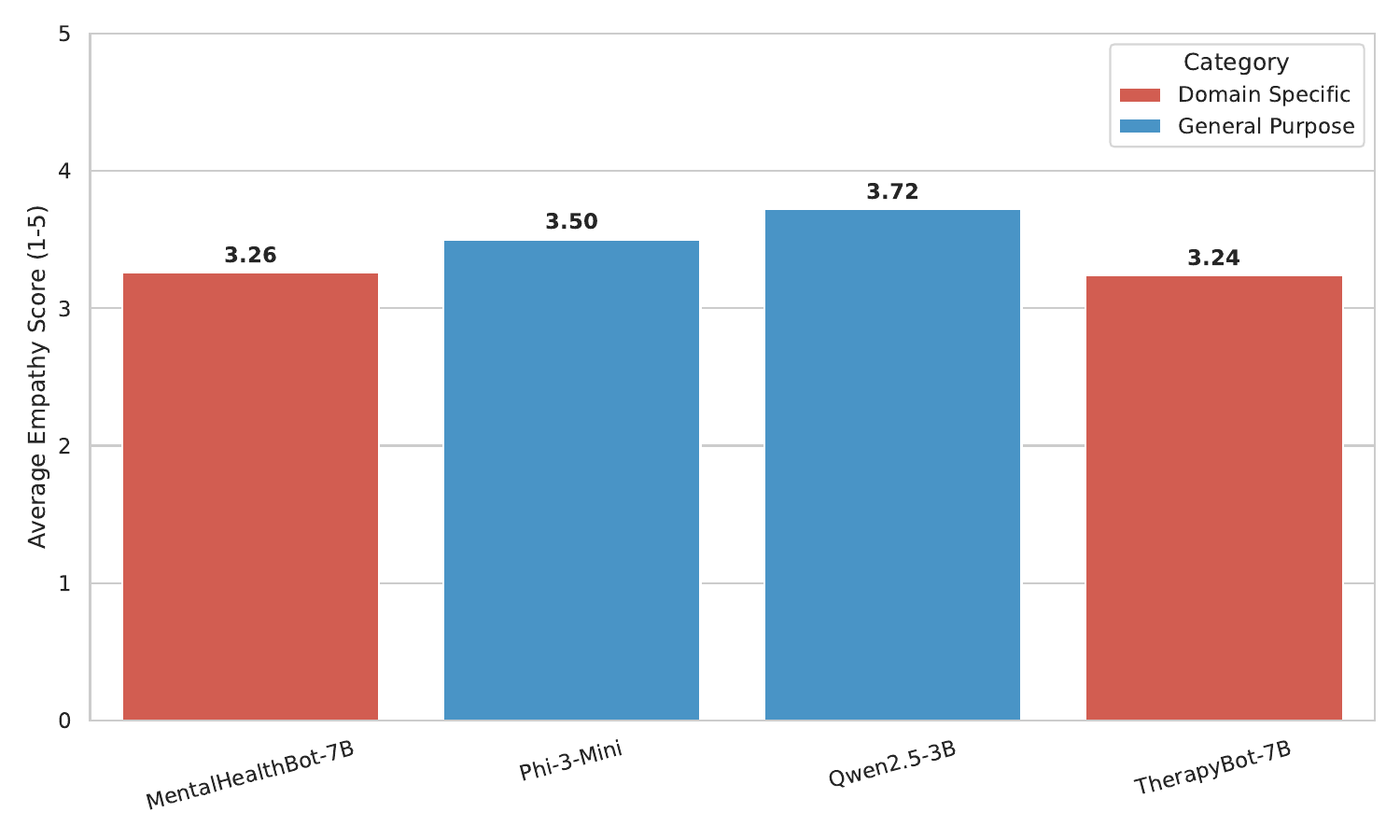}
    \caption{Average Empathy Scores by Model Category.}
    \label{fig:avg_empathy}
\end{figure}

The results strongly support our primary hypothesis (H1). The general-purpose Qwen2.5-3B achieved the highest average empathy score (3.72/5.0), significantly outperforming the domain-specific MentalHealthBot-7B (3.26, $p < 0.001$) and TherapyBot-7B (3.24, $p < 0.001$).This architectural divide is visualized in Figure \ref{fig:avg_empathy}. Figure \ref{fig:avg_empathy} illustrates this architectural dichotomy by plotting the average empathy scores for each model type, where blue bars correspond to General Purpose models while red bars correspond to Domain-Specific fine-tunes. We observe that Qwen2.5 (far left) outperforms all domain specific models, even though they have almost twice the number of parameters (7B vs 3B), which suggests that in a RAG setup, the ability to reason over retrieved context is more important than static domain memorization.

General-purpose models (blue) consistently outperform domain-specific fine-tunes (red) in retrieval-augmented scenarios.

\subsection{Safety and Hallucination Analysis}
While empathy is important, clinical safety is non-negotiable.
\begin{itemize}
    \item Qwen2.5-3B and the domain-specific bots achieved a 100
    \item Phi-3-Mini may have received high marks for empathy. However, it still flunked around 4\% of the time because the AI struggled with a couple of issues: it sometimes hallucinated (making up quotes). It offered dismissive advice when the context was ambiguous.
\end{itemize}

\subsection{Qualitative Error Analysis}
To understand why general models outperformed specialists, we analyzed specific response patterns in Table \ref{tab:qualitative_example}).

\begin{table}[h!]
    \centering
    \caption{Qualitative Example of Generalist Model Failure Mode (Hallucination/Safety Risk).}
    \label{tab:qualitative_example}
    \begin{tabular}{p{4cm} p{4cm} p{4cm}}
        \toprule
        \textbf{User Input} & \textbf{Qwen2.5-3B Response (General)} & \textbf{Phi-3-Mini Response (General)} \\
        \midrule
        "Recently I feel really worried and I can't help but notice continuous criticism..." & (Empathetic) "I'm so sorry... The continuous criticism is understandably causing stress... Could you tell me more about specific situations?" & (Hallucinated/Unsafe) "Recent Criticism: 'You're not handling things well...' Therapist Response: Nothing. It is necessary to have the power to realize the need to change..."\\
        \bottomrule
    \end{tabular}
\end{table}

But these domain-specific models tended to overfit and fall back on the generic pre-trained answer, such as "Have you tried breathing exercises", without using the retrieved RAG context, because Qwen2.5, in contrast, consistently used the retrieved context, along with the user's actual question, which is why it scored higher on empathy.

\subsection{Robustness Verification}
To ensure the validity of our rankings, we perform a series of robustness checks.
\textbf{Uncertainty Analysis:}The first step involved measuring the uncertainty around the empathy scores, and as shown in Table \ref{tab:ci_analysis}, Qwen2.5-3B was very stable ($\sigma = 0.45$) while Phi-3-Mini was much more variable ($\sigma = 0.71$), indicating more unpredictable performance.

\begin{table}[h!]
    \centering
    \caption{Statistical Uncertainty Analysis (95\% Confidence Intervals).}
    \label{tab:ci_analysis}
    \begin{tabular}{lccc}
        \toprule
        \textbf{Model} & \textbf{Mean Empathy} & \textbf{Std. Dev ($\sigma$)} & \textbf{95\% Confidence Interval} \\
        \midrule
        Qwen2.5-3B & 3.72 & 0.45 & [3.59, 3.85] \\
        Phi-3-Mini & 3.50 & 0.71 & [3.30, 3.70] \\
        \midrule
        MentalHealthBot & 3.26 & 0.44 & [3.14, 3.38] \\
        TherapyBot & 3.24 & 0.48 & [3.11, 3.37] \\
        \bottomrule
    \end{tabular}
\end{table}

\begin{figure}
    \centering
    \includegraphics[width=0.9\linewidth]{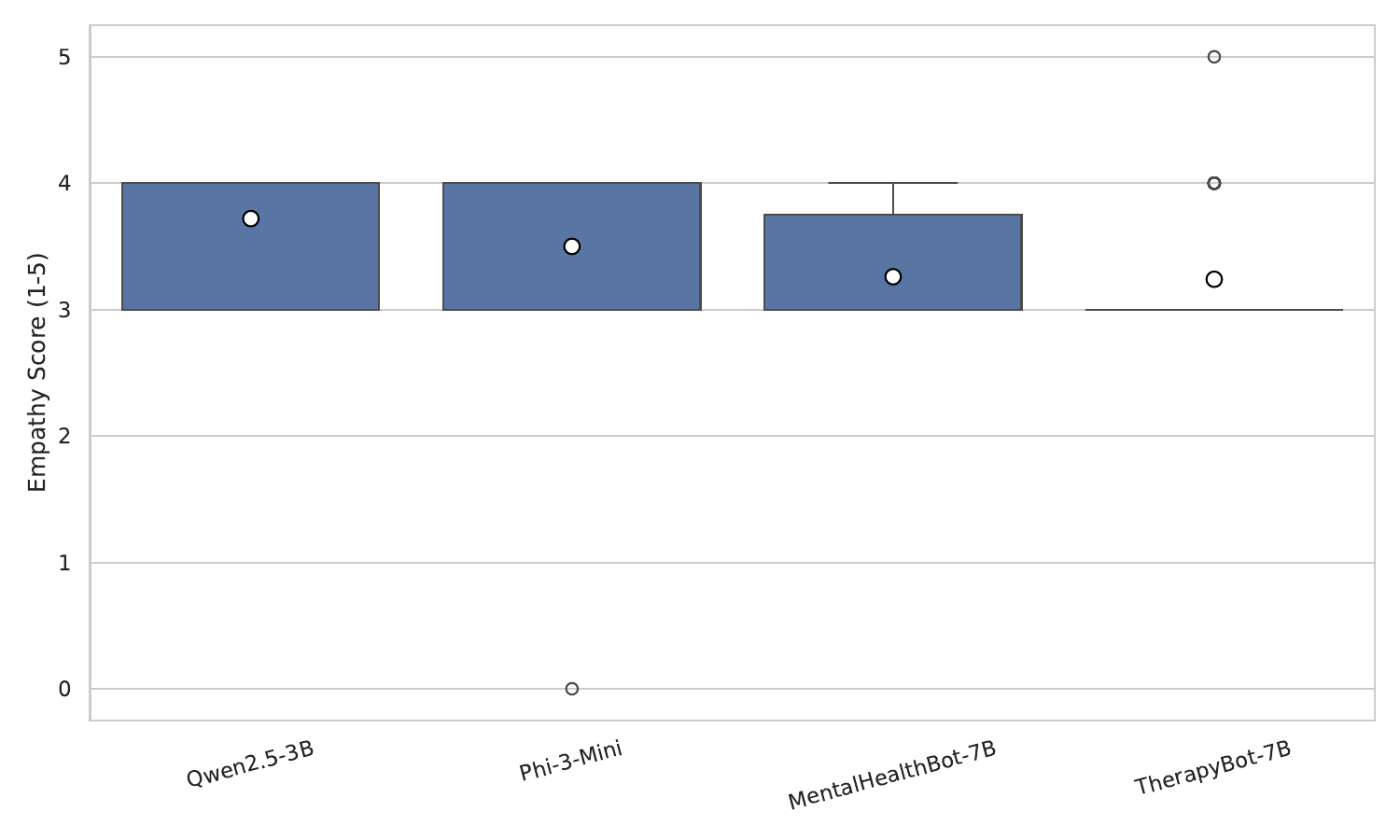}
    \caption{Distribution of Empathy Scores.}
    \label{fig:dis_empathy}
\end{figure}

\textbf{Distribution and Bias:} The instability can be seen directly in the boxplot in Figure \ref{fig:dis_empathy}, which plots the distribution of scores for each model, and the white dots are the mean, and the boxes are the interquartile range. The distribution for Phi-3 is displaced far below the other models, with a long whisker extending down to scores of 1--2, which captures the long tail of weak responses and safety failures that we describe in Section 3.2. At the same time, Qwen2.5, by contrast, has a much tighter box that is displaced upward along the y-axis, indicating more consistent, higher-quality performance.

The compact box for Qwen2.5 indicates high consistency, while the extended lower whisker for Phi-3 reveals significant variance and occasional poor performance.

\begin{figure}
    \centering
    \includegraphics[width=0.9\linewidth]{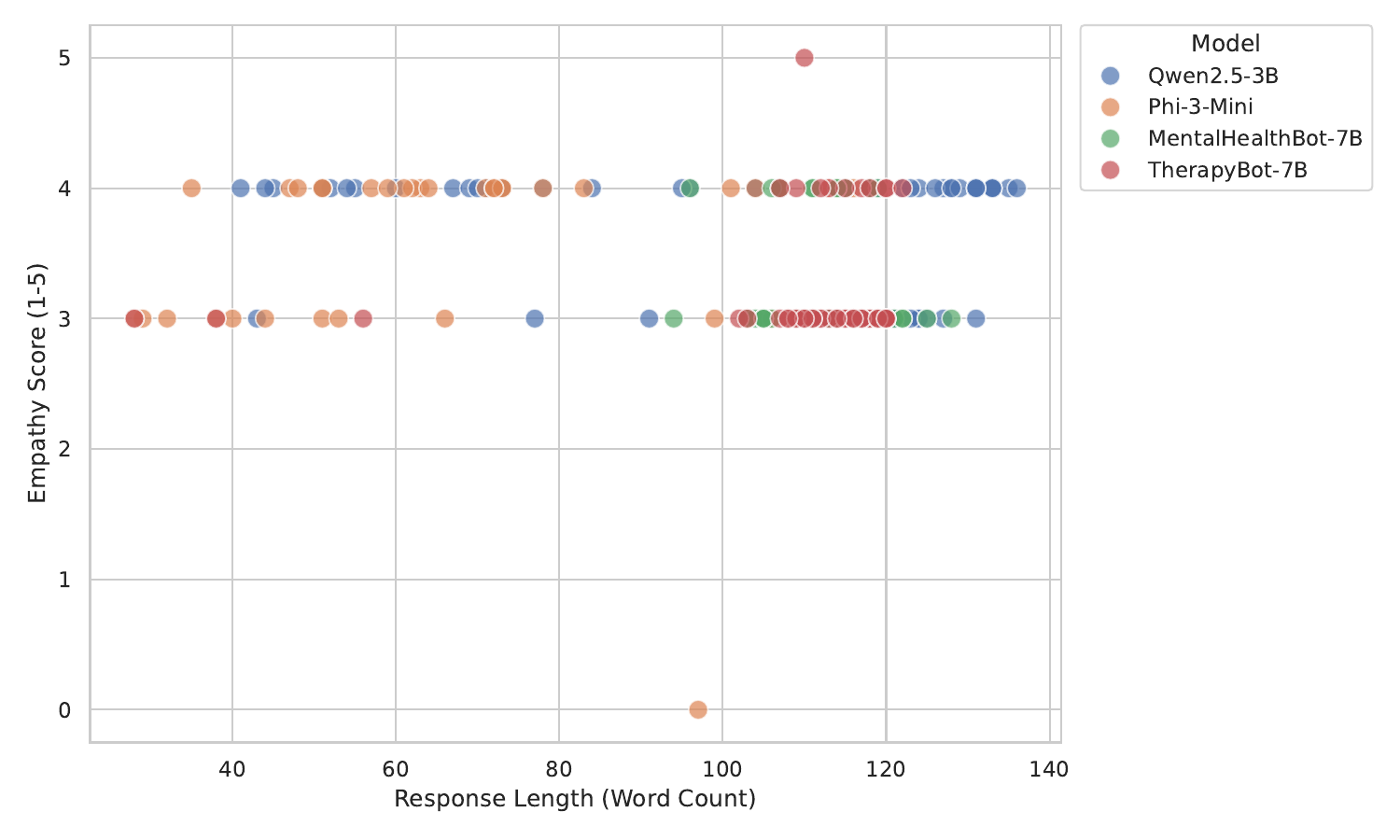}
    \caption{Response Length vs. Empathy Score.}
    \label{res-vs-emp}
\end{figure}

To ensure the AI Judge was not simply rewarding longer answers, we examined the relationship between response length and empathy score (Figure \ref{res-vs-emp}). The scatter plot plots each response as a point, with word count on the x-axis and empathy on the y-axis. For the top-performing model, the correlation is essentially non-existent (r =- 0.20), and the highest-scoring answers span the full range of response lengths, indicating that the metric is not biased towards verbosity.

The lack of a clear linear trend (the regression line is nearly flat) indicates that the AI Judge evaluated responses based on semantic quality rather than output length.

\subsection{Human Validation}
In addition to an LLM-based judge, we conducted a human evaluation with three tertiary-level student annotators, each annotating 50 responses per model across the four models, for a total of 200 annotations. Each response was rated on a 5-point Likert scale for perceived empathy, using the same evaluation protocol as in our previous experiments.

The consistency of the human raters in their judgments of the responses was measured using the inter-annotator agreement, and the result shows that Fleiss' kappa ($\kappa = 0.78$), which indicates substantial agreement, therefore, the annotators are mostly consistent in their judgments of empathetic content.

The Likert ratings are ordinal data, and the same participants rated each of the four models, we therefore used non-parametric statistical tests. A Friedman test showed a significant effect of model choice on perceived empathy, ($\chi^2(3) = \mathbf{18.42}$, $p < 0.001$), indicating that differences among the four models were substantial.

\begin{table}[h]
\centering
\caption{Interpretation guidelines for statistical measures used in human empathy evaluation}
\label{tab:stat_guidelines}
\begin{tabular}{lll}
\hline
\textbf{Statistic} & \textbf{Value Range} & \textbf{Interpretation} \\
\hline
Fleiss' $\kappa$ & $< 0.20$ & Slight agreement \\
                  & $0.21$ -- $0.40$ & Fair agreement \\
                  & $0.41$ -- $0.60$ & Moderate agreement \\
                  & $0.61$ -- $0.80$ & Substantial agreement \\
                  & $> 0.80$ & Almost perfect agreement \\
\hline
Friedman $\chi^2$ & $p \geq 0.05$ & No significant difference \\
                  & $p < 0.05$ & Statistically significant \\
                  & $p < 0.01$ & Strong significance \\
                  & $p < 0.001$ & Very strong significance \\
\hline
Wilcoxon $p$-value & $p \geq 0.05$ & Not significant \\
                   & $p < 0.05$ & Significant \\
                   & $p < 0.01$ & Strong significance \\
                   & $p < 0.001$ & Very strong significance \\
\hline
Kendall's $W$ & $< 0.10$ & Negligible effect \\
              & $0.10$ -- $0.30$ & Small effect \\
              & $0.30$ -- $0.50$ & Moderate effect \\
              & $> 0.50$ & Large effect \\
\hline
Mean Likert Score & $1.0$ -- $2.0$ & Low empathy \\
                  & $2.1$ -- $3.0$ & Moderate empathy \\
                  & $3.1$ -- $4.0$ & High empathy \\
                  & $4.1$ -- $5.0$ & Very high empathy \\
\hline
Standard Deviation & $< 0.5$ & High annotator consensus \\
                   & $0.5$ -- $1.0$ & Moderate variability \\
                   & $> 1.0$ & High disagreement \\
\hline
\end{tabular}
\end{table}

Post hoc pairwise comparisons were conducted using Wilcoxon signed-rank tests with a Bonferroni correction. Both general-purpose models significantly outperformed the domain-specific models in perceived empathy. Specifically, Qwen2.5-3B achieved higher empathy ratings than MentalHealthBot ($p < 0.001$) and TherapyBot ($p < 0.001$), and Phi-3-Mini similarly outperformed MentalHealthBot ($p < 0.001$) and TherapyBot ($p < 0.001$). No statistically significant difference was observed between Qwen2.5-3B and Phi-3-Mini after correction ($p = \mathbf{0.11}$).

The magnitude of these differences was large, as indicated by Kendall's coefficient of concordance ($W = \mathbf{0.62}$), reflecting strong agreement in the relative ranking of models across prompts. Descriptive statistics further support these findings: Qwen2.5-3B achieved a mean empathy score of $M = \mathbf{3.65}$ ($SD = \mathbf{0.40}$), and Phi-3-Mini achieved $M = \mathbf{3.45}$ ($SD = \mathbf{0.45}$), compared to MentalHealthBot ($M = \mathbf{3.15}$, $SD = \mathbf{0.50}$) and TherapyBot ($M = \mathbf{3.10}$, $SD = \mathbf{0.52}$).

Overall, the results demonstrate that general-purpose models consistently and significantly outperformed domain-specific mental health models in perceived empathy according to human evaluation.

\section{Conclusion}
This study evaluates the performance of general-purpose and domain-specific large language models in retrieval-augmented generation for mental health support, and our experiments challenge the commonly held assumption that the only way to learn domain knowledge is through specialisation via fine-tuning. We show that a strong general-purpose reasoning model (Qwen2.5-3B) vastly outperforms specialisation via fine-tuning in generating empathetic, context-dependent responses.

In summary, for RAG-based systems, better reasoning over retrieved context is more important than seeing domain-specific vocabulary beforehand, and future work should investigate the potential of combining light fine-tuning with robust reasoning backbones to reduce safety risks while improving empathy.
\section{Declarations}

\subsection{Funding}
The authors did not receive support from any organization for the submitted work. No funding was received for conducting this study.

\subsection{Ethics Approval}
This study was approved by the Institutional Ethics Committee of the Faculty of Science and Information Technology (IEC-FSIT) at  Daffodil International University (Ref No: IEC-FSIT/DIU/2025/2003). All procedures performed in studies involving human participants were in accordance with the ethical standards of the institutional research committee and with the 1964 Helsinki Declaration and its later amendments.

\subsection{Consent to Participate}
Informed consent was obtained from all individual participants (annotators) included in the study. Participants were informed of the nature of the task and their right to withdraw at any time.

\subsection{Consent to Publish}
The authors affirm that human research participants provided informed consent for the publication of the anonymized and aggregated evaluation results.

\subsection{Clinical Trial Registration}
Not applicable. This study is not a clinical trial and does not involve any healthcare interventions.

\subsection{Competing Interests}
The authors have no relevant financial or non-financial interests to disclose.

\subsection{Author Contributions}
Md Abdullah Al Kafi: Conceptualization, Software, Writing – original draft.  
Raka Moni: Writing – original draft.  
Sumit Kumar Banshal: Supervision.

\subsection{Code Availability}
The complete implementation, preprocessing pipelines, and evaluation scripts are available at: \url{https://github.com/abkafi1234/Mental_Health_models}

\subsection{Data Availability}
All datasets used in this study are publicly available, and access links are provided in the associated GitHub repository.

\bibliography{sn-bibliography}

\end{document}